\documentclass[letterpaper, 10 pt, conference]{ieeeconf}  

\IEEEoverridecommandlockouts                              
\usepackage[utf8]{inputenc}
\usepackage[T1]{fontenc}
\usepackage{amsmath}
\usepackage{graphicx}
\usepackage{amssymb}
\usepackage{comment}
\usepackage{tabularx,caption}
\usepackage{hhline}

\title{\LARGE \bf
Manipulability optimization for multi-arm teleoperation
}

\author{Florian Kennel-Maushart, Roi Poranne, Stelian Coros
}

\newcommand{\x}{\mathbf{x}}
\newcommand{\q}{\mathbf{q}}
\newcommand{\werr}{w_{\text{err}}}
\newcommand{\wreg}{w_{\text{reg}}}
\newcommand{\K}{\mathcal{K}}

\begin{document}

\maketitle
\thispagestyle{empty}
\pagestyle{empty}

\begin{abstract}

Teleoperation provides a way for human operators to guide robots in situations where full autonomy is challenging or where direct human intervention is required. It can also be an important tool to teach robots in order to achieve autonomous behaviour later on.
The increased availability of collaborative robot arms \emph{and} Virtual Reality (VR) devices, provides ample opportunity for development of novel teleoperation methods.
Since robot arms are often kinematically different from human arms, mapping human motions to a robot in real-time is not trivial.
Additionally, a human operator might steer the robot arm toward singularities or its workspace limits, which can lead to undesirable behaviour.
This is further accentuated for the orchestration of \emph{multiple} robots.
In this paper, we present a VR interface targeted to multi-arm payload manipulation, which can closely match real-time input motion.
Allowing the user to manipulate the payload rather than mapping their motions to individual arms we are able to simultaneously guide multiple collaborative arms.
By releasing a \emph{single} rotational degree of freedom, and by using a \emph{local} optimization method, we can improve each arm's manipulability index, which in turn lets us avoid kinematic singularities and workspace limitations. We apply our approach to predefined trajectories as well as real-time teleoperation on different robot arms and compare performance in terms of end effector position error and relevant joint motion metrics.
\end{abstract}

\section{Introduction}

The field of teleoperation of robots and robot arms has seen a lot of activity since both collaborative robot arms and 6 degrees-of-freedom (DOF) input devices have become more affordable and more widely available \cite{lee2020}.
It has been shown that it is generally more intuitive, faster and less mentally exhausting for a human operator to operate a robot arm via head and hand tracking devices rather than via a touch interface, mouse or joystick \cite{rakita2017}.
As the kinematic structure of a robot arm is often different from that of a human arm, and because mostly only the hand position of the operator is tracked rather than every joint of the arm, it is not straightforward to find an end-effector pose that matches the positioning of the arm and the intent behind the operator's movement.
Furthermore, naïve end-effector pose matching can lead to singular or close to singular robot arm configurations which can lead to dangerously fast joint movements for small changes in the end-effector pose or can get the robot stuck in singular configurations, unable to reliably continue to match the operator's pose.

This problem gets accentuated in multi-robot settings, where not one single arm has to match an end-effector pose but instead multiple arms need to support and move a payload in a way that is either given by a predefined trajectory or by real-time human input.
Teleoperation of such a system can exhibit non-intuitive workspace constraints as well as additional singular configurations which arise from the system acting like a closed-loop kinematic chain.
In order for the operator to focus on high-level tasks it is important that the multi-robot system can reliably avoid these configurations and show a graceful degradation of pose-matching reliability, giving the operator the possibility to adapt rather than abruptly getting stuck when approaching singularities and workspace limits.

We first provide a short overview of related work in the field. We then introduce an improvement to traditional Inverse Kinematic (IK) solvers in which we relax one single rotational DOF and provide the solver with a prior configuration that locally maximises the manipulability index of each individual robot arm. This improves the system's ability to avoid dangerous discontinuities in joint configuration which can arise from near-singular configurations.

We then introduce the setup and how we are using our intuitive VR interface for payload manipulation. The interface lets the operator move freely around the virtual robots and modify the payload's pose by simply grasping and moving it. It therefore provides the operator with the ability to safely operate a group of robots remotely, or use it as a simulation tool for training or pre-defining tasks.

Finally, we show the results of applying our method to a group of 3 UR5 robots and 2 ABB YuMis respectively and provide an outlook on future developments.

\begin{figure}[t] 
 \centering
 \includegraphics[width=0.9\linewidth]{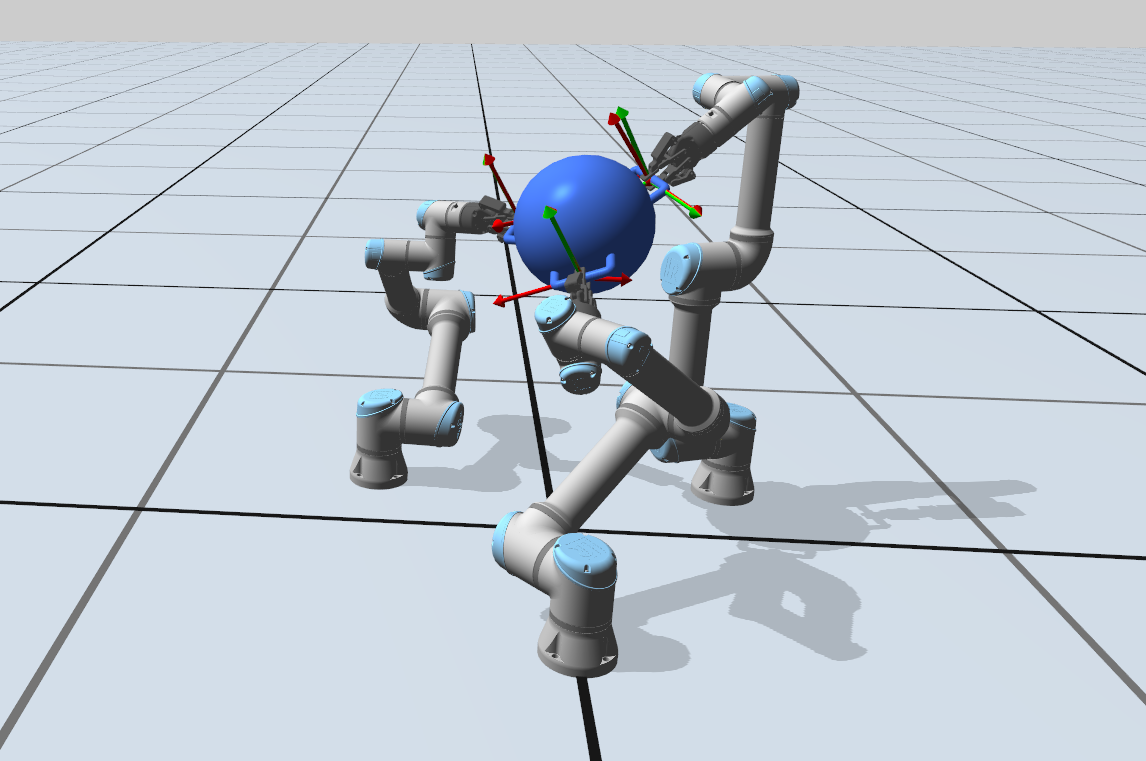} 
 \caption{The setup with 3 UR5 robot arms}
 \label{fig:system}
\end{figure}

\section{Related Work}

In order to quantify the performance of a solution to the inverse kinematic problem, several measures have been proposed, such as the \emph{manipulability index} \cite{yoshikawa2020}, the \emph{condition number} \cite{salisbury1982} and many derivations and adaptions thereof (see e.g. \cite{patel2015} for an overview).
With the appearance of collaborative robots, the interest to operate robots in real-time and potentially close to humans has increased.
In combination with the availability of accurate and affordable 6DOF input devices, the field of teleoperation has seen a lot of exciting developments, drawing inspiration from motion capturing and animation techniques traditionally used for animated movies.
Although the single-arm case has been relatively well studied, dual- and multi-arm collaborative tasks have seen less activity, despite their vast potential for applications in construction, assembly or mobile multi-robot systems.
In the following, we review some of the closely related topics.

\subsection{Manipulability Index and Singularity Avoidance}

 The capability of a robot arm to avoid singular configurations is especially important in the field of physical human-robot interaction (HRI), where a human operator, who cannot easily recognize potentially dangerous configurations, guides the robot e.g. during kinesthetic teaching.
 Several frameworks have been proposed to steer the operator away from singular configurations using for example asymmetric damped least squares \cite{carmichael2017} or admittance control via virtual forces \cite{dimeas2016}.
 These methods often use a \emph{manipulability} or conditioning measure to scale the generated forces depending on the proximity to a singularity and, in the asymmetric case, the sign of the \emph{gradient} of the manipulability.
 Since the absolute distance to the singular configuration is often not required, scaling factors and thresholds which are tuned to work well in HRI applications are introduced.
 A similar method of optimising a measure called \textit{parameter of singularity} in order to avoid singularities was used in \cite{franks2008}.
 To still match the required end-effector trajectory as closely as possible, the authors consider the rotational axis of the tool that is held by the robot arm to be functionally redundant and thus have one redundant degree of freedom to optimise over.
 A similar idea can be found in \cite{zimmermann2020}, where the optimal grasp position on a wooden block is chosen by parametrizing the gripper orientation and position on the block in the presence of obstacles, although the gripper position is not changed anymore once the block has been picked up.
 A method that introduces the manipulability index as an optimization objective into a kinematic task was introduced in \cite{dufour2020}, where the authors showed an improved manipulability index for a predefined trajectory at the cost of reduced trajectory tracking.
 They additionally combined this with obstacle avoidance to show that they are able to maintain a safe distance even with the additional objective.

\subsection{Teleoperation}

With recent advances in Virtual and Augmented Reality devices, virtual teleoperation of robots in 3D space has become more precise and much easier to implement.
Headsets such as the Microsoft Hololens 2 and Oculus Quest 2 are pushing to improve 6DOF head- and hand-tracking and several other headsets such as the HTC Vive are providing affordable and easy to use 6DOF controllers.
Since robots and robot arms are often kinematically different from their human operators, and since, with few exceptions, there is little to no haptic feedback for operators, several methods for intuitive mapping of operator motion and intent onto robotic systems have been proposed.
In \cite{lee2020}, Lee et al. introduced a method of unimanual and bimanual teleoperation of a robot arm, using Oculus Rift IR LED sensors and touch controllers
They showed that their method leads to subjective and objective improvements over traditional kinesthetic teaching methods on moderately challenging tasks.
Rakita et al. \cite{rakita2017} introduced a trade-off between IK and different goals such as obstacle or singularity avoidance.
This allows them to place more importance on avoiding bad configurations during faster and larger hand motions, while giving the operator more precise control over the end-effector pose when approaching or manipulating different objects.
They used HTC Vive controllers as the input device and show improvements over other input methods, such as a 6DOF stylus on several tasks of varying difficulty.

\subsection{Multi-arm manipulation}

While multi-arm coordination generally introduces an additional level of complexity by adding the need for re-grasping or limiting the available workspace if the robot arms are fixed in place, it also provides several benefits such as the modularity, redundancy and increased payload capabilities, while keeping each individual robot relatively simple and cost-efficient.
The growing availability of collaborative robot arms such as Universal Robots' UR5 and UR10 or the ABB YuMi promotes an increasing incentive to build teams of multiple, ready-made robots for tasks that one single robot cannot solve.
In \cite{xian2017}, the authors showed a method that enables the manipulation of large objects by considering two arms as a closed-chain system and introducing \emph{essentially mutually disconnected} components, which allow them to switch between different configurations via re-grasping operations. A method of coupling two robot arms via a virtual object is proposed in \cite{salehian2016}.
The paper focuses on the choice of a grasping position and the interaction of the system with a payload that is handed over by a human collaborator.
Another method that coordinates 4 robot arms by introducing a virtual manipulator is shown in \cite{dehio2018}. The virtual manipulator lets the authors tune the impedance of the system, making it more robust to disturbances such as added payload. 

\begin{figure}[b] 
 \centering
 \includegraphics[width=0.9\linewidth]{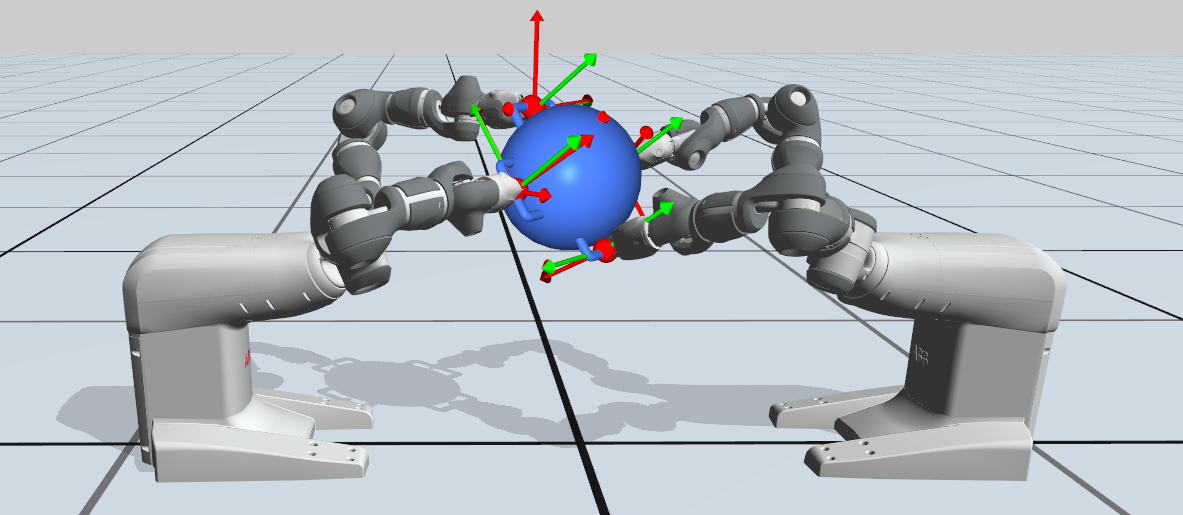} 
 \caption{The setup with 2 ABB YuMi robot arms}
 \label{fig:systemYumi}
\end{figure}

\section{Method}

For many installation and construction tasks the robot arm's end-effector does not actually have to be in a precise location or orientation, as long as it can still guarantee the correct positioning and orientation of the payload.
Examples of this are the installation of glass panels using suction cups, or the positioning and fastening of pipes and tubes to the ceiling on construction sites.
Construction workers can adjust their grasp around the handles or pipes during installation in order to optimize leverage and accessibility.
We propose a method that optimises the manipulation index for each robot arm independently by utilizing one rotational degree of freedom for the optimisation, while limiting the maximum deviation from the fully constrained configuration in order to avoid losing grip or intersecting with the payload. Because the payload is handled by multiple robots simultaneously, its pose is always fully defined, even when opening up the rotational DOF on the robot arm. 

Contrary to previous methods which consider manipulability as part of a global optimization problem (\cite{dufour2020}) or only use it to detect proximity to kinematic singularities (\cite{carmichael2017},\cite{dimeas2016}), we suggest an extremely simple algorithm that improves manipulability while not sacrificing performance and tracking fidelity.
\\
\textbf{Optimization based IK.}
\label{subsec:opt}
We begin by formulating an optimization based IK problem as follows
\begin{equation}\label{eq:IK}
\begin{aligned}
\min_{\q} \quad & \werr\|\K(\q)-x\|^2 + \wreg\|\q-\q_0\|^2\\
\textrm{s.t.} \quad & \q_{\text{min}} < \q < \q_{\text{max}},
\end{aligned}
\end{equation}
where $\q$ are the stacked joint angles, $\q_{\text{min}}$ and $\q_{\text{max}}$ are the joints upper and lower limits, $\K(\q)$ is the forward kinematics function for the pose of the robot's end effector, $\x$ is the target end effector pose, and $\q_0$ is predefined rest pose, which is used as regularization.

We solve this optimization problem using Newton's method.
Joint limits are handled as soft bound constraints using a barrier function
As an initial guess, we provide the solver with the previously computed optimal joint angles, which were based on the previous end effector target.
In the next section, we discuss a local optimization procedure that improve the manipulability of the individual arms.
\\
\textbf{Local manipulability optimization.}
Once \ref{eq:IK} is solved, we can still improve the manipulabilty of of the arms.
The manipulability index is defined by 
\begin{equation}
m(\mathbf{q}) = \sqrt{\det(\mathbf{J}\mathbf{J}^T)}
\end{equation}
Our goal is to find a better orientation of the end effector w.r.t. a single rotation axis, as defined by the handles grasped by the end effector.
To this end, we could potentially solve another optimization problem using a gradient based method.
However, similarly to \cite{dimeas2016} can consider small modification to the joint angles, that, to first order, preserve the pose, except around this one rotation.
This is done via the generalized inverse of the velocity Jacobian,
\begin{equation}
\mathbf{J}^+ = \left(\mathbf{J}^T\mathbf{J}\right)^{-1}\mathbf{J}^T
\end{equation}
In order to compute the first order angle modification we use
\begin{equation}
\mathbf{q'_\pm} = \mathbf{q_0} \pm \mathbf{J^+}\mathbf{M} \Delta t
\end{equation}
where $M$ is a mask matrix (i.e. containing only zeros and ones) that selects the available DOF, and $\Delta t$ is a scaling factor.
In our case, that would be the 4th entry, which corresponds to rotation around the local $x$-axis.

In each iteration of our algorithm, we use these new joint angles $\mathbf{q'_+}$ and $\mathbf{q'_-}$ to compute the new manipulability index for each one of them.
If one of them is higher than the current manipulability we use the configuration  as the new initial guess for the IK solver at the next iteration. If the increase in manipulability falls below a threshold value $\theta_m$ we discard the result in order to avoid oscillating around a locally optimal result and use the current configuration as the initial guess instead.

\section{Experimental Setup}

\subsection{Robots and Payload}
As our method does not depend on the specific structure of the robot arm, it can be applied to a wide variety of different types and groups with different numbers of robots. We validate our method on a setup of 3 6DOF UR5 robots arms and 2 7DOF ABB YuMis respectively due to their wide use in the robotics community and industry and their difference in DOFs as well as number of end effectors per robot. The UR5 setup is shown in Fig. \ref{fig:system}, while the YuMi setup is shown in Fig. \ref{fig:systemYumi}.

\subsection{Tasks}

We test our method on 2 predefined geometric tasks as well as real-time teleoperation scenarios. The geometric tasks consist of square and circular trajectories. The teleoperation tasks are carried out on a Oculus Quest VR headset, using the accompanying 6DOF controllers. In order to ensure repeatability with different configurations we prerecord a user-generated trajectory which we then stream to the controller for subsequent experiments. As proposed in \cite{rakita2018} we compare the positional errors, manipulability index and the joint velocities, accelerations and jerk over the different trajectories. \\

We compare all tasks on both the UR5 and the YuMi setups, using 3 different scenarios:
\begin{itemize}
	\item [A)] IK with full constraints on position and orientation of the end effectors
	\item [B)] IK with masked y and z orientations, allowing free rotation around the x axis
	\item [C)] IK with masked y and z orientations, with our algorithm enabled
\end{itemize}

\subsection{Setup}

\begin{figure}[t] 
 \centering
 \includegraphics[width=0.98\linewidth]{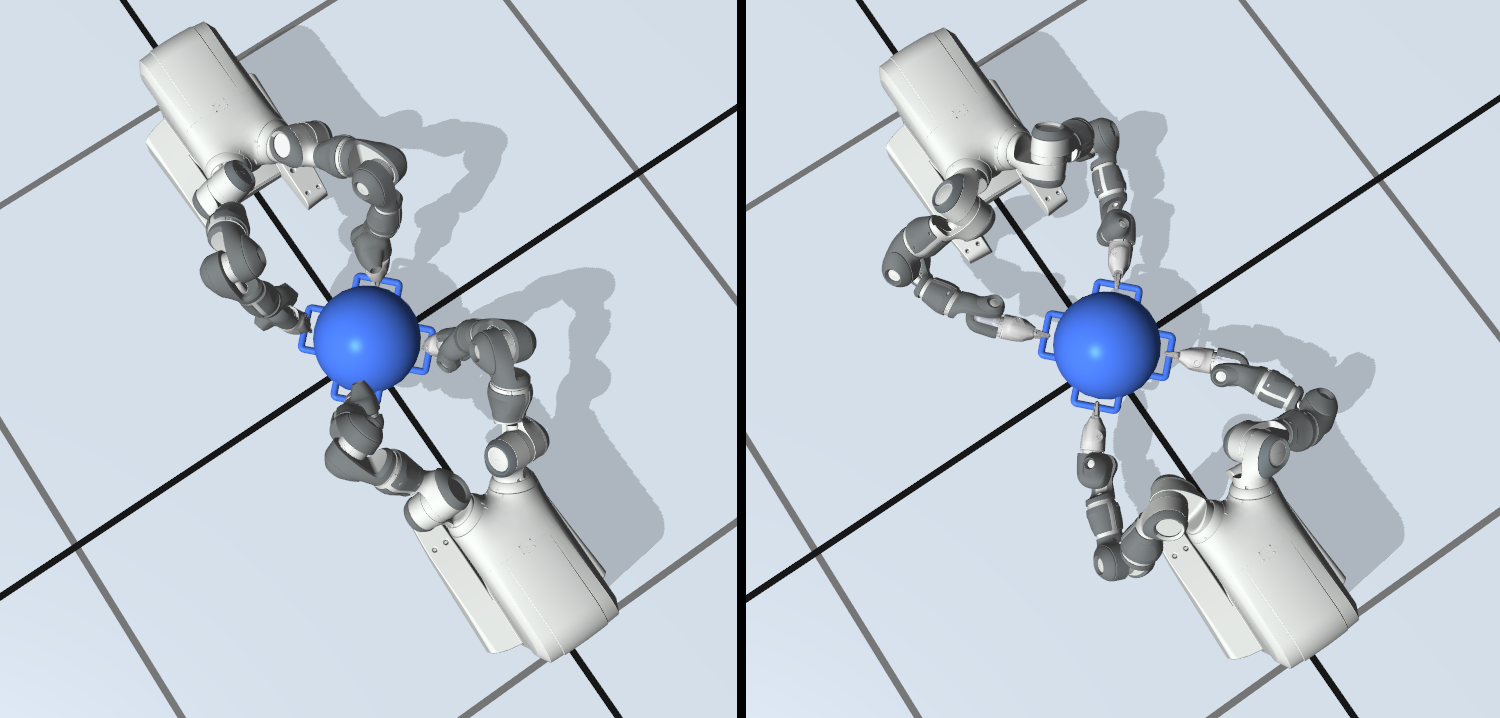} 
 \caption{Comparison of the initial pose of the YuMi arms before (left) and after (right) improving manipulability via rotation around the handle x axis.}
 \label{fig:yumi_manip}
\end{figure}

The payload is represented by a sphere of diameter 0.3m, to which one handle per robot end effector is attached at the equator of the sphere. The handles are initially facing their respective robot arms and are parallel to the floor, and the handle bar has a distance of 5cm from the payload surface. For all experiments the UR5 robot arms are placed in a circle of 1.5m diameter, facing the center of the circle. The YuMis are placed 1.2m apart, facing the center position.To ensure that the experiments are comparable, the weights for our energy function are the same for different trajectories and the different scenarios, namely $w_{\text{err}}=1000$, $w_{\text{reg}}=0.01$ and $w_{\text{lim}}=10\,000$. For our Newton minimizer we are using a maximum of 10 steps and a residual value of $10^{-5}$. For scenario B, we nullify the penalty for rotations around the x axis, as described in Section \ref{subsec:opt}. For scenario C, we additionally enable the manipulability optimization, using a $\Delta t = 0.007$ and a threshold $\theta_m = 0.0001$ for the UR5 experiments and a $\Delta t = 0.005$ and $\theta_m = 0.001$ for the YuMi experiments respectively. These values have been chosen empirically to avoid oscillations while still providing a sufficiently fast convergence towards a better manipulability value. An example of the initial setup for the 2 YuMi robots and the payload can be found in Fig. \ref{fig:yumi_manip}.

\begin{figure}[b] 
 \centering
 \includegraphics[width=\linewidth]{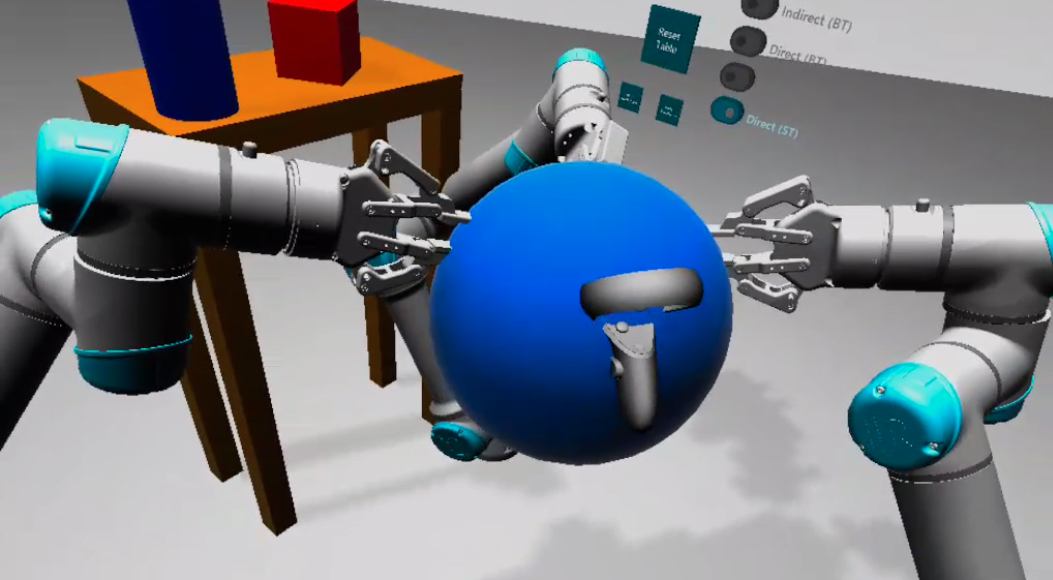} 
 \caption{User interface inside the Oculus Quest. The payload pose can be modified via simple grasping with the controller.}
 \label{fig:vr}
\end{figure}

\subsection{System}

All experiments were run on a Windows 10 machine with an Intel Core i7-9750H CPU @ 2.60GHz, 32GB RAM and an Nvidia Geforce RTX 2080 Max-Q GPU. \\
For the VR teleoperation we used an Oculus Quest 128GB headset and the accompanying 6DOF controller. The VR interface was built in Unity3D and a simple UDP implementation was used to wirelessly transmit manipulation data to the Windows machine. A typical view for an operator inside the headset is shown in Fig. \ref{fig:vr}.

\section{RESULTS and DISCUSSION}

\subsection{Circle trajectory}

\begin{figure}[t] 
 \centering
 \includegraphics[width=0.705\linewidth]{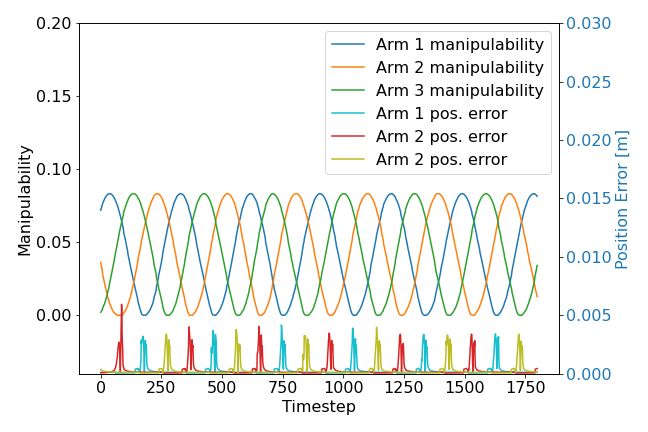} 
 \caption{Scenario A: Comparison of manipulability and position error for the circle trajectory on the 3 UR5 robot arms.}
 \label{fig:manipVSerr}
\end{figure}

The payload is initialized in the middle between all robots. It then moves in positive x direction for 0.2m and finally moves in circles around the origin for a predefined number of steps. If they don't get stuck in a local optimum or because of a singular configuration the arms are theoretically able to meet the end effector goal pose for any point on the trajectory. For each experiment, the position error, velocity, acceleration and jerk values as well as the manipulability were measured over several iterations of the trajectory. The average and standard deviation for the different metrics are summarized in table \ref{t1}. The results indicate that both in scenario B and C the manipulability is improved substantially over scenario A. As shown in the accompanying video, the fully constrained IK solver is not able to continuously match the goal pose and exhibits discontinuities in joint configuration at several points in time. This coincides with the manipulability index approaching a zero value and leads to higher positional error for the end effectors, as can be seen in Fig. \ref{fig:manipVSerr}. While the mean manipulability and position error values as well as the standard deviations are slightly improved when using our method with the UR5 arms, acceleration and jerk are equal or even slightly worse, but generally in the same order of magnitude. Similar results can be seen on the YuMi. Although here we are able to improve the manipulability index as well, we have substantially higher positional errors in general and also when using our method, which might mostly be attributed to the difficulties of the YuMi to still reach far away points on the trajectory. In turn we can see a more substantial improvement in mean velocity, acceleration and jerk.

\subsection{Square trajectory}

\begin{figure}[t] 
 \centering
 \includegraphics[width=\linewidth]{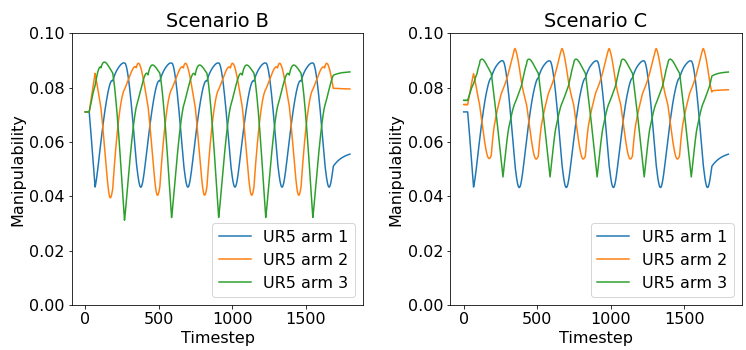} 
 \caption{Qualitative manipulability comparison for scenarios B and C of the square trajectory on the 3 UR5 robot arms}
 \label{fig:squareTraj}
\end{figure}

Similar observations as for the circle trajectory can be made for the square trajectory. The problems of fully constraining the end effector pose are even more obvious here, as the abrupt changes in direction often lead to very high tracking errors, especially when they coincide with low manipulability. While for the UR5 the positional tracking is slightly improved when using our method, the YuMi benefits from lower velocity, acceleration and jerk values, although here improvements are not quite as substantial. This indicates that the sharp changes in velocity don't allow our algorithm to sufficiently quickly improve the manipulability. Therefore, while we achieve slightly higher values for the manipulability index, they do not translate as well to improvements in tracking as for the previous trajectory. Qualitative examples of the UR5's manipulability evolution are shown in Fig. \ref{fig:squareTraj} and the results are summarized in table \ref{t2}.

\begin{table*}[ht]
\centering
\begin{tabular}{|c||c|c|c|c|c|}
	\hline
	 & Pos. [mm]  & Vel. $[\text{rad}/s]$ & Acc. $[\text{rad}/s^2]$& Jerk $[10^{-3}\text{rad}/s^3]$ & $m(\mathbf{q})$ \\
	\hhline{|=#=|=|=|=|=|}
	UR5 Scenario A & $0.32\pm 0.64$  & $0.12\pm 0.93$ &  $0.037\pm 0.006$ & $0.94\pm 1.7$ & $0.043 \pm 0.03$ \\
	\hline
	UR5 Scenario B & $0.025\pm 0.015$ & $0.072\pm 0.03$ & $0.02\pm 0.003$ & $0.53\pm 0.81$ & $0.071 \pm 0.016$\\
	\hline
	UR5 Scenario C & $0.022\pm 0.013$ & $0.071\pm 0.02$ & $0.02\pm 0.003$ & $0.58\pm 0.89$ & $0.074 \pm 0.012$\\
	\hhline{|=#=|=|=|=|=|}
	YuMi Scenario A & $26.9 \pm 48$ & $0.178 \pm 0.178$ & $0.004 \pm 0.008$ & $0.72 \pm 1.4$ & $0.022 \pm 0.008$\\
	\hline
	YuMi Scenario B & $6.9 \pm 13.4$ & $0.143 \pm 0.065$ & $0.003 \pm 0.004$ & $0.58 \pm 0.89$ & $0.025 \pm 0.007$\\
	\hline
	YuMi Scenario C & $8.8 \pm 16.6$ & $0.036 \pm 0.052$ & $0.001 \pm 0.004$ & $0.10 \pm 0.52$ & $0.030 \pm 0.007$\\
	\hline
\end{tabular}
\captionof{table}{Mean results for the circle trajectory}
\label{t1}
\end{table*}

\begin{table*}[ht]
\centering
\begin{tabular}{|c||c|c|c|c|c|}
\hline
	 & Pos. [mm]  & Vel. $[\text{rad}/s]$ & Acc. $[\text{rad}/s^2]$& Jerk $[10^{-3}\text{rad}/s^3]$ & $m(\mathbf{q})$ \\
	\hhline{|=#=|=|=|=|=|}
	UR5 Scenario A & $2.3\pm 8.3$  & $0.12\pm 0.13$ &  $0.0021\pm 0.01$ & $0.33\pm 1.9$ & $0.044 \pm 0.032$ \\
	\hline
	UR5 Scenario B & $0.026\pm 0.014$ & $0.077\pm 0.03$ & $0.0037\pm 0.002$ & $0.07\pm 0.42$ & $0.071 \pm 0.016$\\
	\hline
	UR5 Scenario C & $0.024\pm 0.012$ & $0.078\pm 0.03$ & $0.0055\pm 0.002$ & $0.13\pm 0.56$ & $0.074 \pm 0.014$\\
	\hhline{|=#=|=|=|=|=|}
	YuMi Scenario A & $52.5 \pm 108.7$ & $0.18 \pm 0.48$ & $0.007 \pm 0.062$ & $1.5 \pm 12$ & $0.022 \pm 0.008$\\
	\hline
	YuMi Scenario B & $8.0 \pm 16.0$ & $0.16 \pm 0.13$ & $0.0025 \pm 0.008$ & $0.53 \pm 1.5$ & $0.025 \pm 0.009$\\
	\hline
	YuMi Scenario C & $9.2 \pm 19.1$ & $0.05 \pm 0.14$ & $0.0019 \pm 0.014$ & $0.27 \pm 1.6$ & $0.028 \pm 0.008$\\
	\hline
\end{tabular}
\captionof{table}{Mean results for the square trajectory}
\label{t2}
\end{table*}

\subsection{Teleoperation trajectory}

\begin{figure}[t] 
 \centering
 \includegraphics[width=\linewidth]{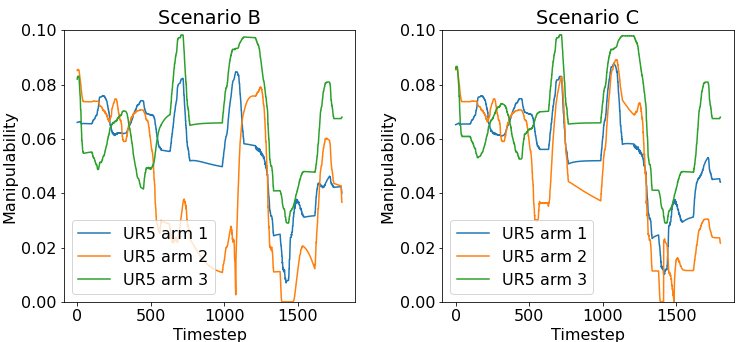} 
 \caption{Qualitative manipulability comparison for scenarios B and C of the teleoperation trajectory on the 3 UR5 robots}
 \label{fig:vrTraj}
\end{figure}

\begin{figure}[b] 
 \centering
 \includegraphics[width=\linewidth]{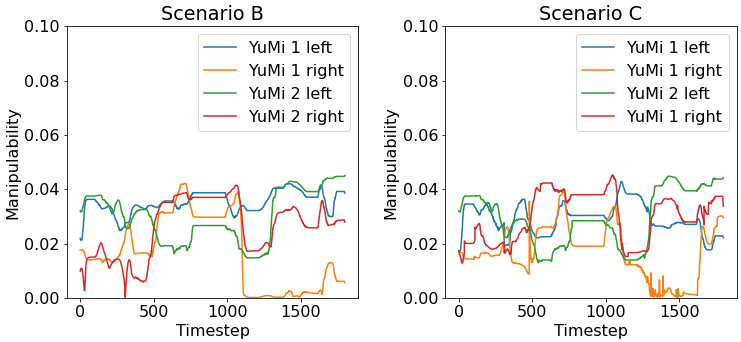} 
 \caption{Qualitative manipulability comparison for scenarios B and C of the teleoperation trajectory on the 2 YuMi robots}
 \label{fig:vrTrajYuMi}
\end{figure}

For the teleoperation trajectory, all parameters were kept the same as for the square and and circular trajectories. The trajectory was pre-recorded and then played back for the different scenarios. The results are summarized in table \ref{t3}. In this scenario, our method clearly performs best, improving the position errors for both the UR5 and the YuMi setup and leading to substantial improvements for the other metrics on the YuMi, while only exhibiting an outlier on the jerk values for the UR5. We believe that this is due to the natural movement of the user, pausing in-between different interactions with the payload, which gives our algorithm time to best adjust the manipulability index. We can also find an explanation for the outliers in Fig. \ref{fig:vrTraj}: The manipulability of arm 2 steadily declines in scenario B, whereas our method is able to make a small adjustment that improves manipulability around timestep 550 and consequently brings the arm into a better configuration. Later on, around timestep 1500, the user guides the payload in a way that makes arm 2 approach its workspace limit, sending the manipulability towards 0 very quickly. Although our method is not able to completely avoid the decline, it nevertheless minimizes the amount of time spent in the unfavorable regime, while the traditional method stays stuck until the user sufficiently adjusts the payload position. These constant adjustments are partially reflected in the mean measurements and also lead to the substantially higher standard deviations. For these cases a simple limitiation of the maximum acceleration would probably suffice to substantially improve the measured values for our method. Fig. \ref{fig:vrTrajYuMi} shows a similar situation for the teleoperation trajectory with the 2 YuMis.

\begin{table*}[ht]
\centering
\begin{tabular}{|c||c|c|c|c|c|}
	\hline
	 & Pos. [mm]  & Vel. $[\text{rad}/s]$ & Acc. $[\text{rad}/s^2]$& Jerk $[10^{-3}\text{rad}/s^3]$ & $m(\mathbf{q})$ \\
	\hhline{|=#=|=|=|=|=|}
	UR5 Scenario A & $1.8\pm 9.9$  & $0.07\pm 0.12$ &  $0.011\pm 0.017$ & $3.2\pm 4.8$ & $0.03 \pm 0.017$ \\
	\hline
	UR5 Scenario B & $0.97\pm 8.1$  & $0.05\pm 0.07$ &  $0.007\pm 0.01$ & $1.8\pm 2.8$ & $0.054 \pm 0.023$ \\
	\hline
	UR5 Scenario C & $0.3\pm 3.8$  & $0.08\pm 0.6$ &  $0.012\pm 0.12$ & $3.1\pm 30$ & $0.057 \pm 0.021$ \\
	\hhline{|=#=|=|=|=|=|}
	YuMi Scenario A & $27 \pm 46$ & $0.085 \pm 0.155$ & $0.008 \pm 0.015$ & $1.5 \pm 2.5$ & $0.020 \pm 0.010$\\
	\hline
	YuMi Scenario B & $5.4 \pm 16$ & $0.059 \pm 0.077$ & $0.007 \pm 0.009$ & $1.5 \pm 2.1$ & $0.027 \pm 0.012$\\
	\hline
	YuMi Scenario C & $4.9 \pm 17$ & $0.019 \pm 0.077$ & $0.001 \pm 0.009$ & $0.3 \pm 1.7$ & $0.029 \pm 0.010$\\
	\hline
\end{tabular}
\captionof{table}{Mean results for the teleoperation trajectory}
\label{t3}
\end{table*}

\subsection{Computational efficiency}

We only need to calculate the generalized inverse of the Jacobian once per end effector and optimization step. Additionally, the calculations for the resulting joint configurations and the optimized manipulability are not heavy and we therefore expected little impact on the overall performance. Indeed, for the experiments presented we add $0.58$ms $\pm 0.23$ms
to a computation time of around 3ms per frame, which still allows the simulation to mostly run at 144 frames per second. 

\section{CONCLUSION}

We introduced a method and a intuitive VR interface for payload manipulation in multi-arm systems. By opening up a rotational degree of freedom and using first order angle modifications to rotate around the free axis we are able to improve the manipulability index for each arm individually. The method was tested on different robots and and for different trajectories, showing that especially in teleoperation where users are not particularly aware of the system limitations we are able to better avoid singular configurations and, in most cases, improve the system behaviour. By adding further improvements such as limits on the maximum accelerations when reaching workspace limits we believe that our method can easily be further improved. The method is computationally efficient and does generally not require a trade-off against other parameters as is the case in \cite{rakita2018} or \cite{dufour2020}. We therefore believe that this is a promising first step in novel developments for future VR teleoperation for multi-arm systems.
\\
\textbf{Mobile bases}: Although our setup consists of robots which are fixed to the ground this is not a limitation of our system and indeed we can already show that it works for robots on omnidirectional mobile bases by adding the DOFs of the base to the overall system and optimizing over the resulting $\mathbf{q}$ vector. Qualitative results of the experiment can be found in the accompanying video. In future iterations we would also like to extend this to directional robot bases.\\
\textbf{Collision avoidance and re-grasping}: To focus on the method in isolation, many of the details of a complete system have been neglected, e.g. collision avoidance.
Additionaly, sometimes a better configuration for the individual robot arm might exist but is not reachable without violating the end effector constraints.
Allowing re-grasping maneuvers might improve the system even further. \\
\textbf{Dynamics and task allocation}: One point of interest in multi-robot systems is that their combined payload capacity allows them to lift heavier objects, while the potential redundancy of additional robots could be used to allow for dynamic re-grasping or allocation of robots to multiple tasks or objects. We believe that this might be especially interesting for larger building sites or automation in larger factory halls.\\

\addtolength{\textheight}{-12cm}   






\end{document}